\documentclass{article}

\usepackage[preprint]{corl_2026} % Uncomment for pre-prints (e.g., arxiv); This is like ``final'', but will remove the CORL footnote.

\usepackage{amsmath,amssymb}
\usepackage{graphicx}
\usepackage{booktabs}
\usepackage{microtype}
\usepackage{tabularx}
\usepackage{array}
\usepackage{subcaption}
\title{Task-Specialized Language-Conditioned Models \\
for Fine-Grained Robotic Manipulation}

\author{
    Chanho Kim\thanks{Both authors contributed equally to this work.} \\
    Independent Researcher \\
    South Korea \\
    \texttt{hch7111@gmail.com}
    \And
    Soyun Choi\footnotemark[1] \\
    Department of Immersive Media Engineering \\
    Sungkyunkwan University \\
    South Korea \\
    \texttt{sychoi9719@skku.edu}
}

\begin{document}
\maketitle

%===============================================================================

\begin{abstract}
Fine-grained robotic manipulation requires instruction grounding, robust perception, and precise contact control. We describe our submission to the MetaFine Open Competition at CoRL 2026. Five independently trained models use a common architecture with frozen DINOv2 and CLIP encoders, trainable language-conditioned feature modulation, and task-specific action heads. We improve demonstrations through contact-aware grasp filtering, instruction-matched data, and expert correction of insertion failures. At inference, task-dependent strategies include photometric test-time augmentation, stochastic action averaging, and uncertainty-triggered replanning. Our submission achieves an official overall score of \textbf{0.914} and ranks \textbf{third among 15 teams}. We analyze the models and their task-specific behavior.
\end{abstract}


%===============================================================================

\section{Introduction}
\label{sec:intro}

Fine-grained manipulation requires robots to identify instructed object parts and execute precise movements despite viewpoint and lighting changes. MetaFine~\citep{xu2026metafine} tests these abilities through part grasping (T1), mug transport (T2), switch manipulation (T3), block placement (T4), and letter insertion (T5), evaluating understanding, perception, and controlled behavior.

\begin{figure}[t]
\centering
\includegraphics[width=\linewidth]{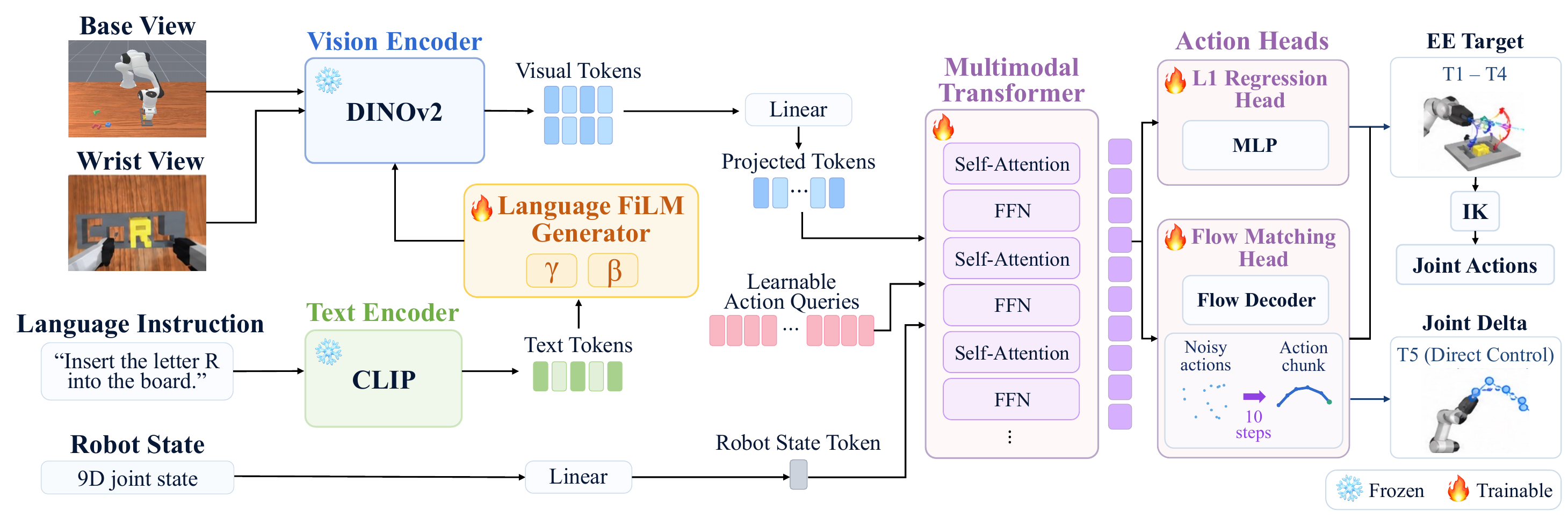}
\caption{\textbf{Overall architecture of our approach.} Frozen DINOv2 and CLIP encoders provide language-conditioned visual features through trainable FiLM modules. A multimodal Transformer drives action heads that produce end-effector targets for T1--T4 or joint-space actions for T5.}
\label{fig:framework}
\end{figure}

We use a shared vision-language architecture with separate task-specific models. To address different failure modes, we combine instruction-conditioned visual features with curated demonstrations and task-dependent execution strategies. Figure~\ref{fig:framework} summarizes the common architecture, while Table~\ref{tab:config} reports the distinct training and inference configurations.

%===============================================================================

\section{Model Architecture and Inference}
\label{sec:method}

\subsection{Language-Conditioned Representation}

Each model receives base and wrist RGB observations, a nine-dimensional robot state containing seven arm joint positions and two gripper-finger positions, and a language instruction. The $512\times512$ images are resized to $224\times224$ and encoded by frozen DINOv2-Base~\citep{oquab2023dinov2}, producing 257 visual tokens per view. A frozen Contrastive Language--Image Pre-training (CLIP) text encoder~\citep{radford2021clip} supplies the language representation. Trainable Feature-wise Linear Modulation (FiLM)~\citep{perez2018film} adjusts 12 visual blocks through instruction-dependent scale and bias, $\widetilde h_\ell=(1+\gamma_\ell(c))\odot h_\ell+\beta_\ell(c)$, with language condition $c$ and identity initialization. The visual features are projected to 512 dimensions and concatenated with a robot-state token and 32 learnable action queries. A four-layer, eight-head Transformer encoder processes these tokens. T1 and T3 use target-aware features, while T5 uses phase conditioning in its decoder.

%===============================================================================

\subsection{Action Prediction and Control}

All models predict 32-step action chunks~\citep{zhao2023act}. T1 and T3 use direct regression with a masked $\ell_1$ loss for target-directed actions. T2, T4, and T5 use conditional flow matching (FM)~\citep{lipman2023flow}, allowing stochastic generation of action sequences. For demonstrated actions $a$, noise $\epsilon\sim\mathcal N(0,I)$, time $t\sim\mathrm{Beta}(1.5,1)$, and Transformer context $c_o$, the velocity field is trained as

\begin{equation}
\label{eq:fm}
\begin{aligned}
x_t &= (1-t)a+t\epsilon,\\
\mathcal L_{\mathrm{FM}} &= \mathbb E\big[\|v_\theta(x_t,t,c_o)-(\epsilon-a)\|_2^2\big].
\end{aligned}
\end{equation}

Here, $t=1$ denotes noise and $t=0$ denotes the demonstrated action. Generation integrates backward from $t=1$ to $t=0$ using ten Euler steps. For T1--T4, the heads output ten-dimensional end-effector (EE) targets comprising three position coordinates, a six-dimensional rotation representation, and a gripper command. Damped-least-squares inverse kinematics (IK) converts the target pose into arm commands, while the gripper command is passed separately. T5 directly predicts seven arm joint deltas and one gripper command.

%===============================================================================

\subsection{Task-Dependent Inference}

For T1 and T3, photometric test-time augmentation (TTA)~\citep{shanmugam2021tta} produces eight action predictions from the original observations and seven brightness- and contrast-perturbed variants. We select the $\ell_1$ medoid, which minimizes the total distance to the other predictions, without updating model parameters. For FM-based models, we instead average eight action chunks generated from different Gaussian noise samples under the same observation to reduce dependence on a single sample.

For T2 and T5, the execution horizon adapts to disagreement among the eight sampled chunks. The action disagreement at future step $j$ is calculated as follows:

\begin{equation}
\label{eq:uncertainty}
u_j=\frac{1}{d_a}\sum_{k=1}^{d_a}\operatorname{Std}_{m=1}^{8}\!\left[a_{j,k}^{(m)}\right].
\end{equation}

Here, $d_a$ is the action dimension and $a_{j,k}^{(m)}$ is the normalized $k$-th action component at step $j$ in sample $m$.
The model executes at least four steps and then truncates the chunk before the first step with $u_j>0.11$ in normalized action units. It then generates a new chunk from a fresh observation. This is related to adaptive action chunking (AAC)~\citep{liang2026aac}, but uses a fixed sample-disagreement threshold rather than entropy-based horizon selection. T1 additionally uses a wrist-conditioned joint residual around gripper transitions to refine contact-sensitive actions.

%===============================================================================

\section{Task-Specific Learning}
\label{sec:training}

We train task-specific models with AdamW, batch size 16, and a learning rate of $10^{-4}$ with 1,000 warm-up steps followed by cosine decay. Camera-view augmentation broadens the demonstrations across viewpoint changes. Table~\ref{tab:config} summarizes the deployed configurations.

\begin{table}[t]
\centering
\caption{Task-specific training and deployment settings. Episode counts include rendered camera views before applying training masks.}
\label{tab:config}
\small
\setlength{\tabcolsep}{3.5pt}
\begin{tabular}{@{}llrll@{}}
\toprule
Task & Head / action & Episodes & Training emphasis & Inference \\
\midrule
T1 & $\ell_1$ / EE & 5,024 & Contact audit, rollouts & TTA, residual \\
T2 & FM / EE & 1,200 & Directional demos & FM mean, replan \\
T3 & $\ell_1$ / EE & 4,800 & Paired target grounding & TTA \\
T4 & FM / EE & 4,644 & Color-conditioned demos & FM mean \\
T5 & FM / joint & 6,723 & DAgger-style correction & FM mean, replan \\
\bottomrule
\end{tabular}
\end{table}

%===============================================================================

\paragraph{T1. Contact-aware part grasping.}
We filter scripted demonstrations to exclude premature contact, arm collisions, and unintended object displacement. After initial training, successful model rollouts replace 30\% of expert body-grasp demonstrations to incorporate states encountered during execution. Grasp-point features condition the action Transformer, while a 3D grasp-vector head provides auxiliary supervision.

\paragraph{T2--T4. Instruction-specific demonstrations.}
T2 uses 1,200 camera-rendered episodes spanning left, right, and forward mug transport. T3 pairs red and blue switch instructions within the same scenes, yielding 4,800 rendered episodes. An instruction-conditioned grounding module converts the correspondence between observed switch attributes and the requested target into a soft action-conditioning token. T4 uses 4,644 rendered episodes for color-conditioned block placement. These datasets emphasize directional, target-selection, and sequential placement requirements, respectively.

\paragraph{T5. Corrective learning for insertion.}
The initial model could grasp and carry letters but often stalled near the narrow slots. We therefore adopt a Dataset Aggregation (DAgger)-style strategy~\citep{ross2011dagger}. Following failed rollouts, an expert resumes from recorded near-slot states to complete insertion. Collection emphasizes letters with higher failure rates and retains 458 successful corrections. Rendering these corrections under three camera conditions and combining them with the original multi-view demonstrations yields 6,723 episodes. The FM decoder also employs phase- and language-conditioned Low-Rank Adaptation (LoRA)~\citep{hu2022lora} with auxiliary slot-alignment supervision.

% %===============================================================================

\section{Experimental Results}
\label{sec:results}

\subsection{Official Evaluation}

The competition reports understanding ($U$), perception ($P$), and controlled behavior ($B$). Perception uses area under the success curve (AUSC) across visual perturbations. Each task score is $S_t=(U_t+P_t+B_t)/3$, and the overall score is $S_{\mathrm{all}}=\frac{1}{5}\sum_{t=1}^{5}S_t$. Our final submission ranks \textbf{third among 15 teams} with the official overall score \textbf{0.914}. Table~\ref{tab:official} reports the task scores.

\begin{table}[!ht]
\centering
\caption{Official MetaFine evaluation results. Scores are rounded for display.}
\label{tab:official}
\small
\setlength{\tabcolsep}{6pt}
\begin{tabular}{@{}lrrrr@{}}
\toprule
Task & $P$ & $U$ & $B$ & Score \\
\midrule
T1 Grasp part & 0.735 & 0.775 & 0.775 & 0.762 \\
T2 Move mug & 0.967 & 1.000 & 1.000 & 0.989 \\
T3 Toggle switch & 0.988 & 1.000 & 1.000 & 0.996 \\
T4 Place blocks & 1.000 & 1.000 & 1.000 & 1.000 \\
T5 Insert letter & 0.816 & 0.825 & 0.825 & 0.822 \\
\midrule
\textbf{Overall} & -- & -- & -- & \textbf{0.914} \\
\bottomrule
\end{tabular}
\end{table}

%===============================================================================

\subsection{Task-wise Results and Analysis}

\paragraph{T1. Grasp part.}

T1 achieves an official score of 0.762, with perception AUSC of 0.735 and understanding of 0.775. Part-level grasping remains less reliable than mug transport, switch manipulation, and block placement under the official evaluation.

\begin{figure}[t]
\centering
\begin{subfigure}[t]{0.48\linewidth}
    \centering
    \includegraphics[width=\linewidth]{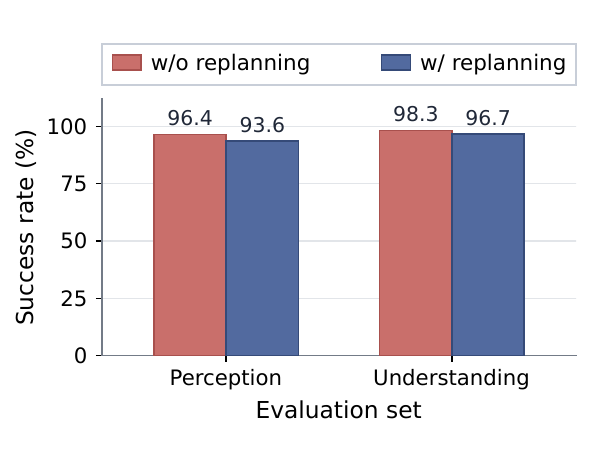}
    \caption{T4 block placement.}
    \label{fig:t4_replanning}
\end{subfigure}
\hfill
\begin{subfigure}[t]{0.48\linewidth}
    \centering
    \includegraphics[width=\linewidth]{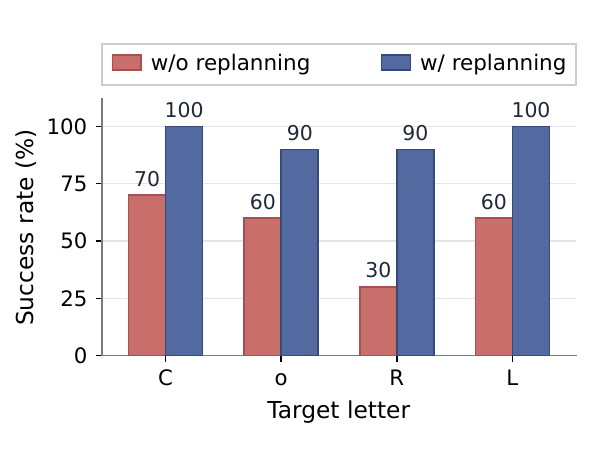}
    \caption{T5 letter insertion.}
    \label{fig:t5_replanning}
\end{subfigure}
\caption{\textbf{Task-dependent effects of replanning.} Episode success rates with and without uncertainty-triggered replanning for (a) T4 block placement and (b) T5 letter insertion.}
\label{fig:task_analysis}
\end{figure}

%===============================================================================

\paragraph{T2. Mug transport.}

T2 achieves 0.989, with perfect understanding and perception AUSC of 0.967. The multi-view-trained model reliably performs left, right, and forward transport under the evaluated visual perturbations.

%===============================================================================

\paragraph{T3. Switch manipulation.}

T3 achieves 0.996, with perfect understanding and perception AUSC of 0.988. The model trained with paired instructions and a grounding token reliably selects the requested switch across the evaluated viewpoint and lighting changes.

%===============================================================================

\paragraph{T4. Block placement.}

T4 achieves 1.000 across all official dimensions. As shown in Fig.~\ref{fig:task_analysis}\subref{fig:t4_replanning}, fixed-horizon execution outperformed replanning with identical weights and eight-sample averaging, achieving 96.4\% versus 93.6\% over 140 perception episodes and 98.3\% versus 96.7\% over 60 understanding episodes. Block placement may require fewer alignment corrections than narrow-slot insertion in T5, while repeated chunk regeneration could disrupt motion continuity. We therefore retained fixed-horizon execution for T4.

%===============================================================================

\paragraph{T5. Letter insertion.}

T5 achieves an official score of 0.822 and 86.9\% understanding success in a separate 320-episode evaluation. Figure~\ref{fig:task_analysis}\subref{fig:t5_replanning} compares execution strategies using identical weights, eight-sample action averaging, and 10 episodes per letter. Replanning increased insertion success from 55\% to 95\%, with R improving from 30\% to 90\%, supporting its use for T5. Remaining failures involved incomplete seating near the target slot.

%===============================================================================

\section{Conclusion}
\label{sec:conclusion}

We presented a shared language-conditioned model architecture with task-specific data refinement and execution strategies to address five distinct manipulation tasks. Our final submission achieves \textbf{0.914}, ranking \textbf{third among 15 teams}. The execution ablations support task-dependent use of replanning rather than applying it uniformly.

%===============================================================================

% \section{Citations}
% \label{sec:citations}

% 	Citations can be made using either \textbackslash citep\{\} or \textbackslash citet\{\}, depending from the appropriateness. To avoid the citation moving to the next line, it is often a good practice to replace the space before with a tilde (\~{}) character.
% 	Example 1: ``CoRL is the best conference ever~\citep{Gauss1857}.''
% 	Example 2: ``\citet{Lagrange1788} proved, both theoretically and numerically, that CoRL is the best conference ever.''
	
%===============================================================================

\clearpage
% The acknowledgments are automatically included only in the final and preprint versions of the paper.

% \acknowledgments{If a paper is accepted, the final camera-ready version will (and probably should) include acknowledgments. All acknowledgments go at the end of the paper, including thanks to reviewers who gave useful comments, to colleagues who contributed to the ideas, and to funding agencies and corporate sponsors that provided financial support.}

%===============================================================================

\bibliography{main}  % .bib

@String(CVPR = {Proc. IEEE/CVF Conf. Comput. Vis. Pattern Recognit.})

@String(ICCV = {Proc. IEEE/CVF Int. Conf. Comput. Vis.})

@String(ICML = {Proc. Int. Conf. Mach. Learn.})

@String(ICLR = {Proc. Int. Conf. Learn. Represent.})

@String(AAAI = {Proc. AAAI Conf. Artif. Intell.})

@String(RSS = {Proc. Robot. Sci. Syst.})

@String(AISTATS = {Proc. Int. Conf. Artif. Intell. Statist.})

@String(TMLR = {Trans. Mach. Learn. Res.})

@article{xu2026metafine,
  title={Beyond Binary Success: A Diagnostic Meta-Evaluation Framework for Fine-Grained Manipulation},
  author={Xu, He-Yang and Zhang, Pengyuan and Ge, Zongyuan and Hao, Xiaoshuai and Belongie, Serge and Geng, Xin and Peng, Yuxin and Wei, Xiu-Shen},
  journal={arXiv preprint arXiv:2605.19986},
  year={2026}
}

@article{oquab2023dinov2,
  title={{DINOv2}: Learning Robust Visual Features without Supervision},
  author={Oquab, Maxime and Darcet, Timoth{\'e}e and Moutakanni, Th{\'e}o and Vo, Huy V. and Szafraniec, Marc and Khalidov, Vasil and Fernandez, Pierre and Haziza, Daniel and Massa, Francisco and El-Nouby, Alaaeldin and Assran, Mido and Ballas, Nicolas and Galuba, Wojciech and Howes, Russell and Huang, Po-Yao and Li, Shang-Wen and Misra, Ishan and Rabbat, Michael and Sharma, Vasu and Synnaeve, Gabriel and Xu, Hu and J{\'e}gou, Herv{\'e} and Mairal, Julien and Labatut, Patrick and Joulin, Armand and Bojanowski, Piotr},
  journal=TMLR,
  year={2024}
}

@inproceedings{radford2021clip,
  title={Learning Transferable Visual Models from Natural Language Supervision},
  author={Radford, Alec and Kim, Jong Wook and Hallacy, Chris and Ramesh, Aditya and Goh, Gabriel and Agarwal, Sandhini and Sastry, Girish and Askell, Amanda and Mishkin, Pamela and Clark, Jack and Krueger, Gretchen and Sutskever, Ilya},
  booktitle=ICML,
  volume={139},
  pages={8748--8763},
  year={2021}
}

@inproceedings{perez2018film,
  title={{FiLM}: Visual Reasoning with a General Conditioning Layer},
  author={Perez, Ethan and Strub, Florian and de Vries, Harm and Dumoulin, Vincent and Courville, Aaron C.},
  booktitle=AAAI,
  pages={3942--3951},
  year={2018}
}

@inproceedings{zhao2023act,
  title={Learning Fine-Grained Bimanual Manipulation with Low-Cost Hardware},
  author={Zhao, Tony Z. and Kumar, Vikash and Levine, Sergey and Finn, Chelsea},
  booktitle=RSS,
  year={2023}
}

@inproceedings{lipman2023flow,
  title={Flow Matching for Generative Modeling},
  author={Lipman, Yaron and Chen, Ricky T. Q. and Ben-Hamu, Heli and Nickel, Maximilian and Le, Matthew},
  booktitle=ICLR,
  year={2023}
}

@inproceedings{shanmugam2021tta,
  title={Better Aggregation in Test-Time Augmentation},
  author={Shanmugam, Divya and Blalock, Davis and Balakrishnan, Guha and Guttag, John},
  booktitle=ICCV,
  pages={1214--1223},
  year={2021}
}

@inproceedings{liang2026aac,
  title={Adaptive Action Chunking at Inference-time for Vision-Language-Action Models},
  author={Liang, Yuanchang and Wang, Xiaobo and Wang, Kai and Wang, Shuo and Peng, Xiaojiang and Chen, Haoyu and Chua, David Kim Huat and Vadakkepat, Prahlad},
  booktitle=CVPR,
  pages={20802--20811},
  year={2026}
}

@inproceedings{ross2011dagger,
  title={A Reduction of Imitation Learning and Structured Prediction to No-Regret Online Learning},
  author={Ross, St{\'e}phane and Gordon, Geoffrey and Bagnell, Drew},
  booktitle=AISTATS,
  volume={15},
  pages={627--635},
  year={2011}
}

@inproceedings{hu2022lora,
  title={{LoRA}: Low-Rank Adaptation of Large Language Models},
  author={Hu, Edward J. and Shen, Yelong and Wallis, Phillip and Allen-Zhu, Zeyuan and Li, Yuanzhi and Wang, Shean and Wang, Lu and Chen, Weizhu},
  booktitle=ICLR,
  year={2022}
}

\end{document}